\documentclass[sigconf]{acmart}
\usepackage{algorithm}
\usepackage{algorithmic}

\usepackage{multirow}
\usepackage{subcaption}
\usepackage{xcolor}

\setcopyright{rightsretained}

\acmDOI{10.1145/3571600.3571634}
\acmArticle{34}
\copyrightyear{2022}
\acmYear{2022}
\setcopyright{acmcopyright}\acmConference[ICVGIP'22]{Proceedings of the Thirteenth Indian Conference on Computer Vision, Graphics and Image Processing}{December 8--10, 2022}{Gandhinagar, India}
\acmBooktitle{Proceedings of the Thirteenth Indian Conference on Computer Vision, Graphics and Image Processing (ICVGIP'22), December 8--10, 2022, Gandhinagar, India}
\acmPrice{15.00}
\acmDOI{10.1145/3571600.3571634}
\acmISBN{978-1-4503-9822-0/22/12}

\begin{document}

\title{Spectrum-inspired Low-light Image Translation for Saliency Detection}

\author{Kitty Varghese$^{1}$, Sudarshan Rajagopalan$^{2}$, Mohit Lamba$^{1}$, Kaushik Mitra$^1$}
\def \authors{Kitty Varghese, Sudarshan Rajagopalan, Mohit Lamba, Kaushik Mitra }
\affiliation{%
\institution{$^1$ Indian Institute of Technology, Madras
\country{India}}
}
\affiliation{%
  \institution{$^2$Madras Institute of Technology
  \country{India}}
}

\renewcommand{\shortauthors}{}

\begin{abstract}
Saliency detection methods are central to several real-world applications such as robot navigation and satellite imagery. 
However, the performance of existing methods deteriorate under low-light conditions because training datasets mostly comprise of well-lit images. 
 One possible solution is to collect a new dataset for low-light conditions. This involves pixel-level annotations, which is not only tedious and time-consuming but also infeasible if a huge training corpus is required. 
We propose a technique that performs classical band-pass filtering in the Fourier space
to transform well-lit images to low-light images and
use them as a proxy for real low-light images.
Unlike popular deep learning approaches which require learning thousands of parameters and enormous amounts of training data, the proposed transformation is fast and simple and easy to extend to other tasks such as low-light depth estimation.

Our experiments show that the state-of-the-art saliency detection and depth estimation networks trained on our proxy low-light images perform significantly better on real low-light images than networks trained using
existing strategies. 
\end{abstract}

\ccsdesc[500]{Computing methodologies~Interest point and salient region detections}

\ccsdesc[500]{Computing methodologies~Interest point and salient region detections}

\keywords{Low-light and salient object detection}
\maketitle
\section{Introduction}

\label{intro}
Saliency detection models aim to identify prominent subjects in a scene, which is useful in several tasks such as robot navigation~\cite{robotic_nagivation,robot2}, satellite imagery~\cite{satellite,satellite2},
video summarization~\cite{video_summarization}, foreground annotation~\cite{foreground_annotation}, and action recognition~\cite{action_recognition,action_recognition2}.
\begin{figure}[!t]
  \centering
  \captionsetup[subfigure]{labelformat=empty}
  \captionsetup[subfigure]{justification=centering}
  \captionsetup[subfigure]{aboveskip=1pt, belowskip=5pt}
    \begin{tabular}{cccc}
    \centering
    \setlength{\tabcolsep}{0pt}
     \subfloat[\scriptsize\textbf{a)}  Real Well-lit ]{\includegraphics[width=.30\linewidth]{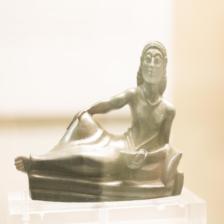}}&\hspace*{-1.5em}
      \subfloat[\scriptsize \textbf{b)} Real low-light ]{\includegraphics[width=.30\linewidth]{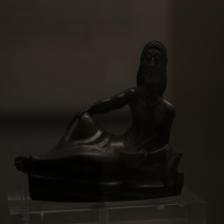}}&\hspace*{-1.5em}
          \subfloat[\scriptsize \textbf{c) Train: }Real well-lit\\  \textbf{Test: }Real well-lit]{\includegraphics[width=.30\linewidth]{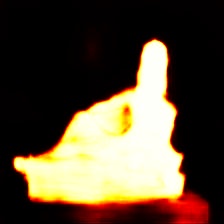}}&\hspace*{-1.5em}
        \\
      \hfill
      \subfloat[\scriptsize\textbf{d) Train: }Real well-lit\\  \textbf{Test: }Real low-light]{\includegraphics[width=.30\linewidth]{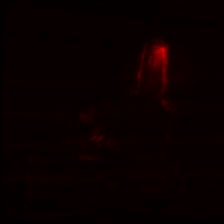}}&\hspace*{-1.5em}
      \subfloat[\scriptsize\textbf{e) Train:} FDA imgs\\  \textbf{Test: }Real low-light]{\includegraphics[width=.30\linewidth]{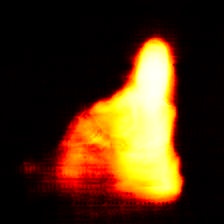}}&\hspace*{-1.5em}
      \subfloat[\scriptsize\textbf{f) Train: }Our imgs\\  \textbf{Test: }Real low-light]{\includegraphics[width=.30\linewidth]{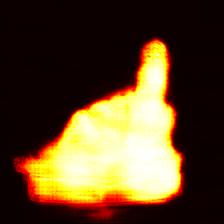}}&\hspace*{-1.2em}
      \multirow[t]{2}{*}{\rotatebox{90}{\subfloat{\includegraphics[width=0.700\linewidth,height=.035\linewidth]{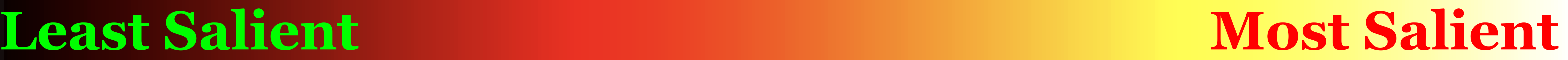}}}}
    \end{tabular}
    \\
    \begin{tabular}{cccc}
    \centering
 \setlength{\tabcolsep}{3pt}
     \subfloat[\scriptsize\textbf{g)} \scriptsize Real well-lit ]{\includegraphics[width=.30\linewidth]{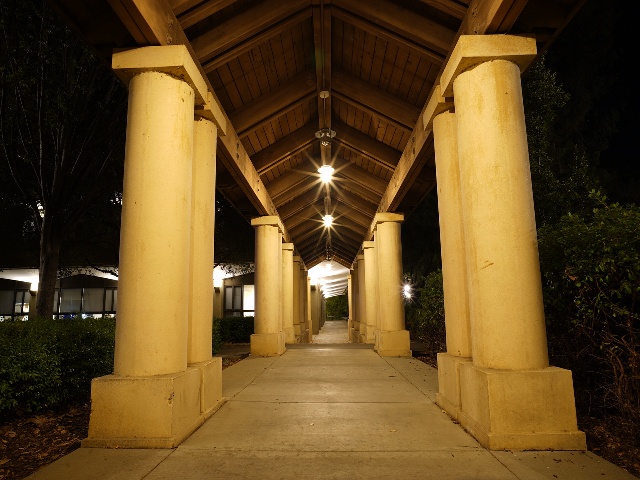}}&\hspace*{-1.5em}
      \subfloat[\scriptsize\textbf{h)} Real low-light]{\includegraphics[width=.30\linewidth]{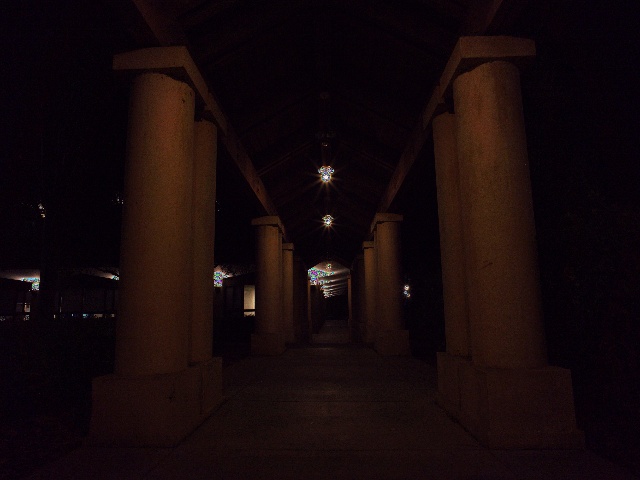}}&\hspace*{-1.5em}
     \subfloat[\scriptsize\textbf{i) Train: }Real well-lit\\  \textbf{Test: }Real well-lit]{\includegraphics[width=.30\linewidth]{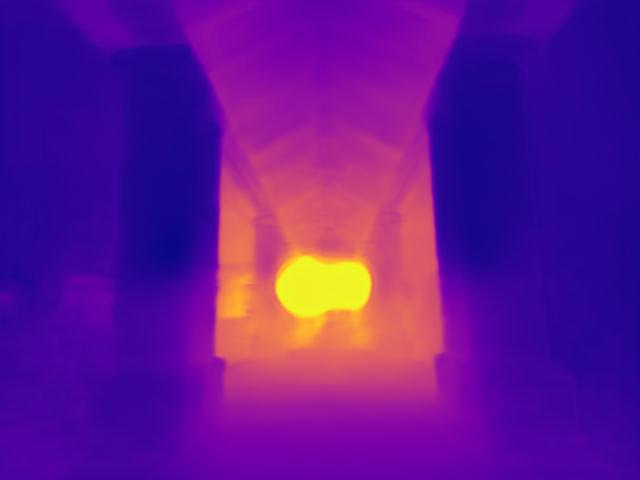}}&\hspace*{-1.5em}\\
      \hfill
      \subfloat[\scriptsize\textbf{j) Train: }Real well-lit\\  \textbf{Test: }Real low-light]{\includegraphics[width=.30\linewidth]{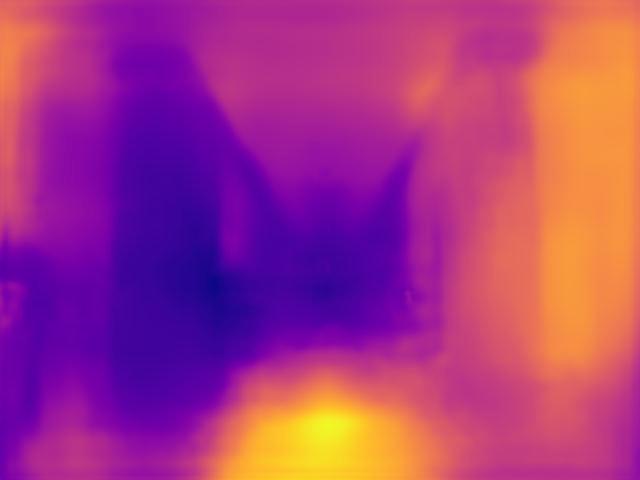}}&\hspace*{-1.5em}
      \subfloat[\scriptsize\textbf{k) Train:} FDA imgs\\  \textbf{Test: }Real low-light]{\includegraphics[width=.30\linewidth]{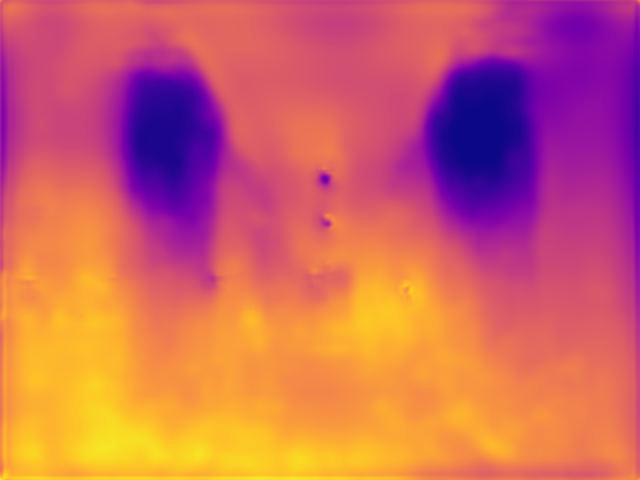}}&\hspace*{-1.5em}
      \subfloat[\scriptsize \textbf{l) Train: }\scriptsize Our imgs\\  \textbf{Test:} Real low-light]{\includegraphics[width=.30\linewidth]{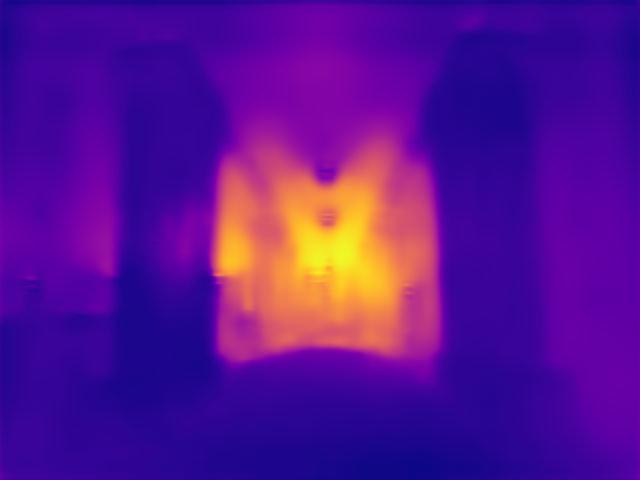}}&\hspace*{-1.2em}
      \multirow[t]{2}{*}{\rotatebox{90}{\subfloat{\includegraphics[width=0.5425\linewidth,height=.035\linewidth]{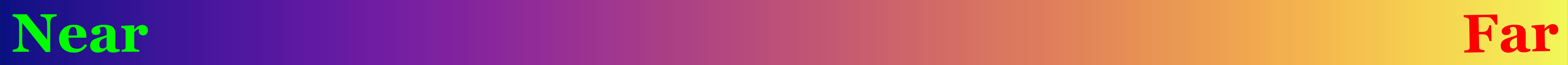}}}}
      \end{tabular}
      \caption{
      Saliency and depth estimation networks perform poorly for low-light images, see \textit{(d)} \& \textit{(j)}, because datasets mainly comprise of well-lit images. We propose a simple transformation from well-lit to low-light images. Training existing models on our proxy low-light images significantly boosts the model's performance on \textit{real} low-light images, see \textit{(f)} \& \textit{(l)}.}
      \label{motivation}
\end{figure}
In a real-world scenario, these applications
require the saliency detection model
to perform well in both good and bad lighting conditions.
But, past studies in this domain~\cite{liu2018picanet,qin2019basnet,gao2020highly} have focused mainly on good lighting conditions with their effectiveness deteriorating for low-light images, as shown in Fig.~\ref{motivation}.

 \begin{figure*}[t!]
    \centering
    \includegraphics[width=0.8\linewidth]{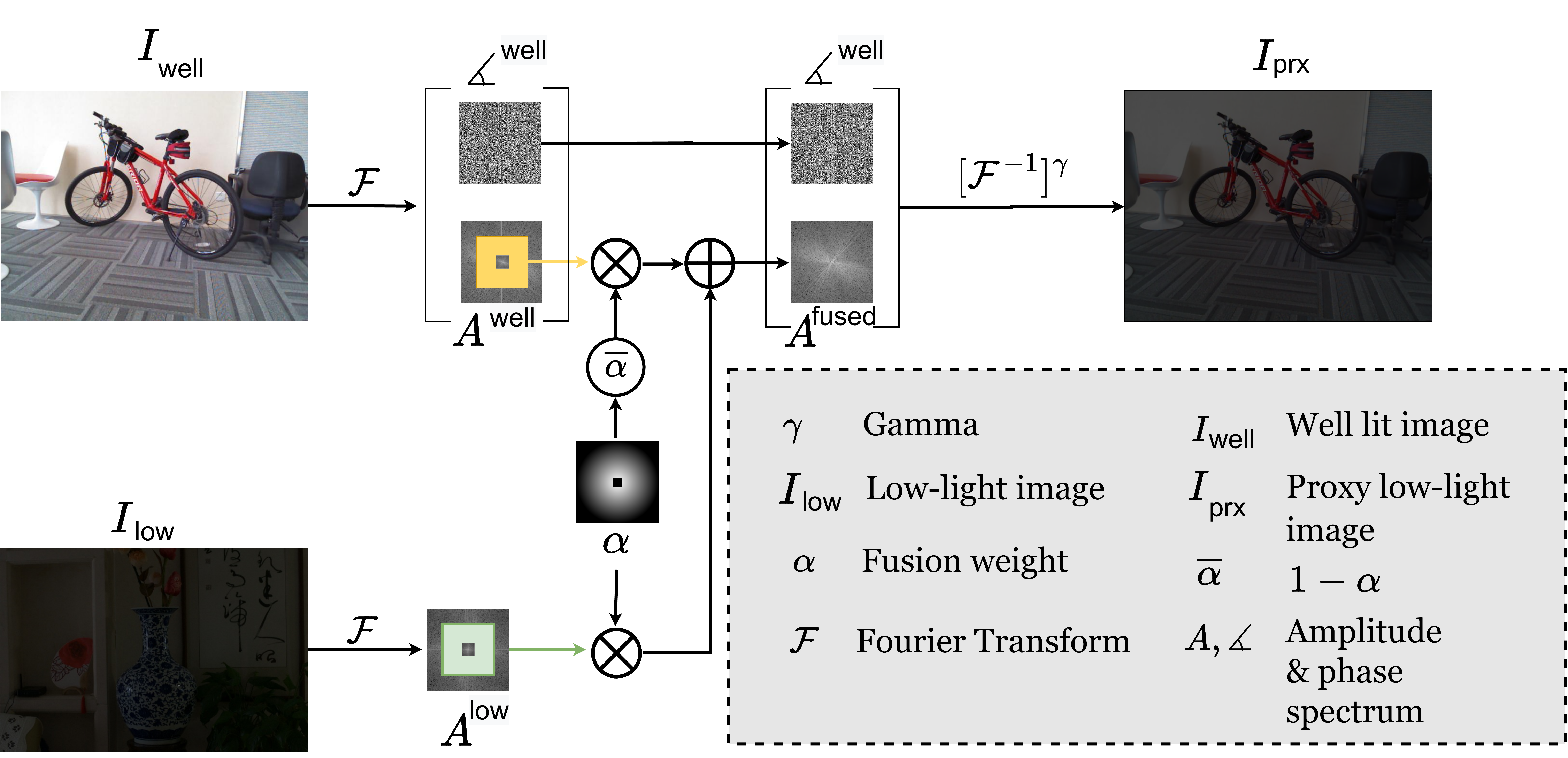}
    \caption{Block diagram of the proposed method.}
    \label{fig:proposed_method}
\end{figure*}
An obvious solution is to pre-process low-light images using existing restoration methods~\cite{guo2020zero,retinex,EnlightenGAN} and then feed them to saliency detection models trained for well-lit images.
But our experiments indicate that this does not yield satisfactory results, see Fig.~\ref{Qualitatitve Analysis}.
Another alternative is to create a new dataset for low-light conditions.
This can be done by manually annotating salient objects in existing low-light datasets~\cite{SID,SICE} or by retouching well-lit saliency detection datasets~\cite{NLPR} in image editing softwares like Adobe Lightroom and GIMP~\cite{wei2018deep,underexposed_cvpr2019}. 
Either-way, this could be laborious, time-consuming, and perhaps even infeasible when a large amount of training data is required.

To alleviate the above challenges, several image translation~\cite{comogan,anokhin2020high,park2020contrastive} and domain adaptation~\cite{yang2020fda,domain_adaptation_1} methods have been proposed.
For example,
HiDT~\cite{anokhin2020high}
adopts an encoder-decoder architecture to decompose a well-lit image into its style and content and consequently uses adversarial learning to transform well-lit images into low-light images.
Nonetheless, such GAN-based solutions are difficult to train and susceptible to problems such as mode collapse~\cite{goodfellow}.
Recently, Yang \textit{et al.}~\cite{yang2020fda} proposed a simple domain adaptation technique, called Fourier Domain Adaptation (FDA), wherein they swap the low frequencies of the source and target domain images. In the present context, source domain represents well-lit images while target domain represents low-light images.
However, FDA is likely to introduce ringing artifacts in the transformed image due to the Gibbs phenomenon~\cite{Oppenheim_book_discrete}, leading to sub-optimal results, as discussed in Sec.~\ref{sec:ablation}. 

To alleviate above problems, we propose a transformation that fuses the amplitude spectrum of a well-lit image with that of a low-light image using band-pass filtering, as shown in Fig.~\ref{fig:proposed_method}. We keep the phase spectrum as it is, because it contains structural information about the source image~\cite{phase_oppenheim}. 
During band-pass filtering of the amplitude spectrum, we also perform a windowing operation to facilitate smooth transition of frequencies and to curb ringing artifacts.
The proxy low-light image is finally obtained by computing the inverse Fourier transform of the fused amplitude response and the phase spectrum of the well-lit image.
These transformed well-lit images into proxy images are then used to train existing networks for real low-light conditions.  Our proposed approach is computationally and memory efficient as it requires tuning a couple of hyper-parameter and needs only $3-4$ real low-light images for the transformation of well-lit images into proxy images. This is in contrast with popular deep-learning-based models which require training hundreds of parameters and a lot of images. 
For the aforementioned reason, our proposed transformation can be easily generalised to other computer vision tasks in low-light conditions. We show that networks trained using our proxy images perform significantly better on real low-light images for downstream computer vision tasks such as saliency prediction and depth estimation.

Our contributions can be summarised as below:
\begin{itemize}
     \item We propose a technique for transforming well-lit images into proxy low-light images, which can then be used to train existing networks for real low-light conditions.
      
     \item Unlike popular deep-learning-based solutions, our  approach  requires  tuning  only  a  couple  of  hyper-parameters and a handful of real low-light images.
     Thus, the proposed transformation can be easily generalized to other computer vision tasks.
     \item We demonstrate both qualitatively and quantitatively that the state-of-the-art saliency detection and depth estimation networks trained on our proxy low-light images perform significantly better on real low-light images.
 \end{itemize}

\section{Related Works}
Saliency prediction models can be classified as bottom-up and top-down models. Bottom up saliency models use low-level features and are stimuli driven as discussed in \cite{itti1998model}. Work by Goferman et al.~\cite{goferman2011context} detects saliency by computing the local and global contrast. Kim et al.~\cite{kim2014salient} in their work used a regression based model and color transform to calculate local and global saliency. These bottom up saliency networks often fail in detecting salient objects when the background is cluttered and in low contrast regions. \\
Whereas, top-down models use high level features to detect salient objects. Xu et al.~\cite{xu2015salient} in their work predict saliency maps using a support vector machine (SVM) model. A covariance based CNN model was used by Mu et al.~\cite{mu2018salient} to learn saliency values in image patches. Dong et al.~\cite{dong2021bcnet} used feature fusion and feature aggregation in their bidirectional collaboration network (BCNet) for detecting salient objects. It is observed that top down saliency networks demand high computational requirements, yet they fail to predict accurate boundaries of salient objects in low-light conditions. Thus, we see that low-light saliency detection is a largely unexplored problem. We propose a method to address this problem by generating proxy low-light images from well-lit images.

Past works have also explored image translation methods to solve similar problems but not saliency detection in low light conditions. We give a brief overview of them. Park et al.~\cite{park2020contrastive}, used unpaired image-to-image translation using contrastive learning for domain adaptation. Anokhin et al.~\cite{anokhin2020high}, used the style and content representation of an image to translate into desired domain. Long et al.~\cite{Image_translation_1} used per-pixel regression for classification to solve image-to-image translation. Li et al.~\cite{Image_translation_2} used PatchGAN architecture to locate style statistics. Isola et al.~\cite{Image_tranlation_3} used Pix2pix to map functions between input and output images. However, most of these methods use deep networks which are data hungry and need a lot of training time. Recently, Yang et al.~\cite{yang2020fda} proposed Fourier domain adaptation (FDA) which overcomes these limitations as they do not need a large training corpus.
\section{Spectrum inspired low-light image translation}
\subsection{Method Overview}
We propose a method to convert well-lit images into proxy images. Our main objective is to reduce the domain gap for downstream computer vision applications by fusing the statistics of low-light and well-lit images. This enables networks to perform downstream vision tasks in low-light conditions even in the absence of real low-light datasets. We do not place much emphasis on making the proxy images look visually indistinct from real low-light images. 

Our method takes inspiration from the fact that in the Fourier representation of an image, it is the phase that carries most relevant information needed to restore the image, and changes made to the amplitude spectrum do not alter higher-level semantics. 
We thus retain the phase spectrum of the well-lit image as it is. The amplitude spectrum of the well-lit image, on the other hand, is fused with the amplitude spectrum of a real low-light image using weighted averaging. Further, to preserve the colors we use band-pass filtering and adopt 2D windowing for suppressing the ringing artifacts. 
Using our method mitigates the problem of building a large real low-light dataset which may be time consuming and laborious. Since, our method mainly involves modification of the spectral characteristics of images, the computation efficiency depends mainly on that of the FFT algorithm. This makes it very fast compared to training neural networks for image translation and has a very low memory footprint (See Sec.~\ref{sec:time}).

\begin{algorithm}[tb]
\caption{Proxy Dataset Generation}
\label{alg:ALG1} 
\begin{flushleft}
\textbf{Input}: $\mathcal{D}_{\textrm{well}}$: dataset of well-lit images; $\mathcal{D}_{\textrm{low}}$: pool of real low-light images. \\
\textbf{Hyperparameters}: $\lambda_l$, $\lambda_u$, $\gamma$.\\
\textbf{Remarks}: $\mathcal{D}_{\textrm{low}}$ can have unpaired images with respect to $\mathcal{D}_{\textrm{well}}$ and should have at least $1$ real low-light image, i.e. $|\mathcal{D}_{\textrm{low}}|\ge1$.\\
\textbf{Output}: $\mathcal{D}_{\textrm{prx}}$: dataset of proxy images.
\end{flushleft}
\begin{algorithmic}[1] 
\STATE $\mathcal{D}_{\textrm{prx}}=\{\}$
\FOR{$I_{\textrm{well}} \textrm{ in } \mathcal{D}_{\textrm{well}}$}
\IF{$|\mathcal{D}_{\textrm{low}}|>1$}
\STATE Sample a real low-light image, i.e. $I_{\textrm{low}} \sim\mathcal{D}_{\textrm{low}}$
\ELSE
\STATE $I_{\textrm{low}}=\mathcal{D}_{\textrm{low}}$
\ENDIF
\STATE $I_{\textrm{low}}=\textrm{resize}(I_{\textrm{low}}, \textrm{size}=\textrm{dim}(I_{\textrm{well}}))$
\STATE $A^{\textrm{well}}$, $\measuredangle^{\textrm{well}}$ = DFT($I_{\textrm{well}}$)
\STATE $A^{\textrm{low}}$, $\measuredangle^{\textrm{low}}$ = DFT($I_{\textrm{low}}$)
\STATE Define $\mathcal{R}=\mathcal{R}_{u}-\mathcal{R}_{l}$ where $\mathcal{R}_{u}, \mathcal{R}_{l}$ are given by Eq.~\ref{eq:regions}
\STATE Compute mask $\alpha_{B}$ as defined in Eq.~\ref{eq:modified}
\STATE $A^{\textrm{fused}} = \alpha_{B} \cdot A^{\textrm{low}} + (1-\alpha_{B}) \cdot A^{\textrm{well}}$
\STATE $I_{\textrm{prx}}$ = $\left [IDFT (A^{\textrm{fused}},\measuredangle^{\textrm{well}})\right]^{\gamma}$
\STATE Append ${I}_{\textrm{prx}}$ to $\mathcal{D}_{\textrm{prx}}$
\ENDFOR
\RETURN{$\mathcal{D}_{\textrm{prx}}$}
\end{algorithmic}
\end{algorithm}

\begin{figure*}
    \centering
    \setlength{\tabcolsep}{1pt}
    \def\arraystretch{1.2}
    \begin{tabular}{cccccccccc}
        &&&\multicolumn{7}{c}{\includegraphics[width=0.7\linewidth, height=.06\columnwidth]{IEEE-SPL-LaTex-1/images/colorbar_hot_new.pdf}}\\
\multirow{2}{*}[20pt]{    \includegraphics[width=0.1\linewidth, height=80pt]{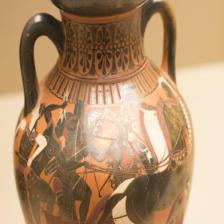}} &  
\multirow{2}{*}[20pt]{
  \includegraphics[width=0.1\linewidth, height=80pt]{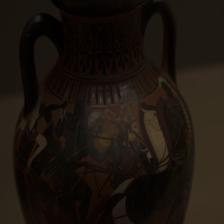}} & \rotatebox{90}{~~~\scriptsize CSNet ~\cite{gao2020highly} } & 
      \includegraphics[width=0.1\linewidth, height=80pt]{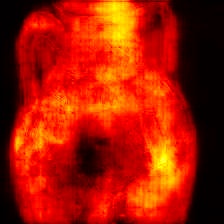} & 
    \includegraphics[width=0.1\linewidth, height=80pt]{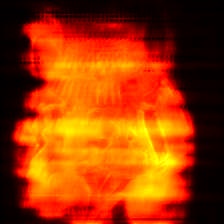} &
    \includegraphics[width=0.1\linewidth, height=80pt]{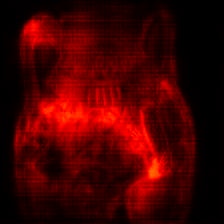} &  
        \includegraphics[width=0.1\linewidth, height=80pt]{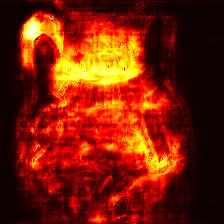} & 
    \includegraphics[width=0.1\linewidth, height=80pt]{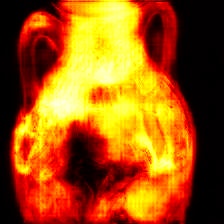}&
    \includegraphics[width=0.1\linewidth, height=80pt]{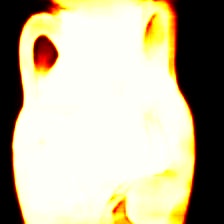} &  
    \includegraphics[width=0.1\linewidth, height=80pt]{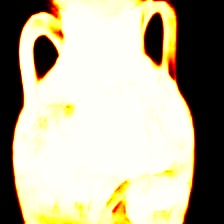}\\
    \medskip
         &  
     &  \rotatebox{90}{~~\scriptsize BASNet~\cite{qin2019basnet}} &
         \includegraphics[width=0.1\linewidth, height=80pt, height=80pt]{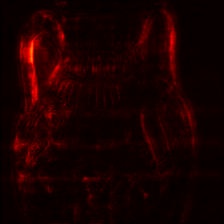} & 
    \includegraphics[width=0.1\linewidth, height=80pt]{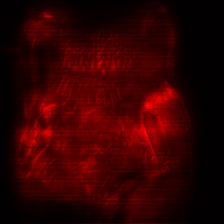} &  
    \includegraphics[width=0.1\linewidth, height=80pt]{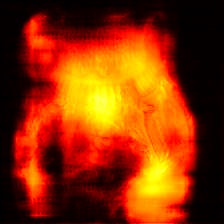} & 
        \includegraphics[width=0.1\linewidth, height=80pt]{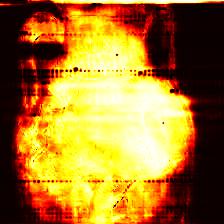} & 
        \includegraphics[width=0.1\linewidth, height=80pt]{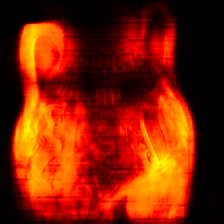} & 
    \includegraphics[width=0.1\linewidth, height=80pt]{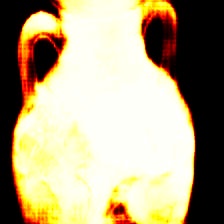} &
\includegraphics[width=0.1\linewidth, height=80pt]{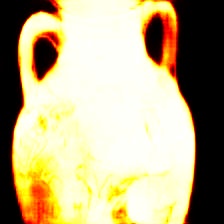}\\
\hline 
\\
  
\multirow{2}{*}[20pt]{    \includegraphics[width=0.1\linewidth, height=80pt]{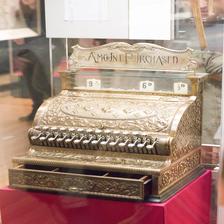}} &  
\multirow{2}{*}[20pt]{
    \includegraphics[width=0.1\linewidth, height=80pt]{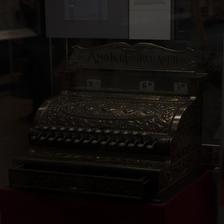}} & \rotatebox{90}{\scriptsize~~~CSNet ~\cite{gao2020highly} } & 
    
          \includegraphics[width=0.1\linewidth, height=80pt]{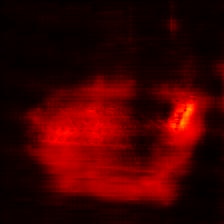}&
        \includegraphics[width=0.1\linewidth, height=80pt]{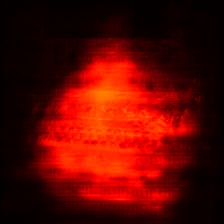} &
    \includegraphics[width=0.1\linewidth, height=80pt]{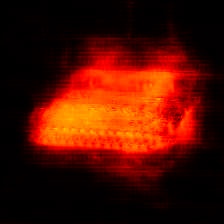} & 
        \includegraphics[width=0.1\linewidth, height=80pt]{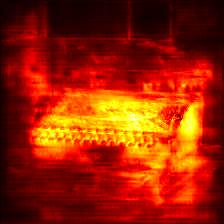} & 
    \includegraphics[width=0.1\linewidth, height=80pt]{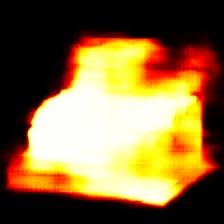}&  
    \includegraphics[width=0.1\linewidth, height=80pt]{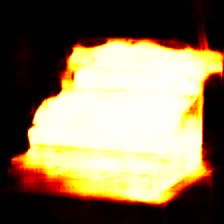} &  
    \includegraphics[width=0.1\linewidth, height=80pt]{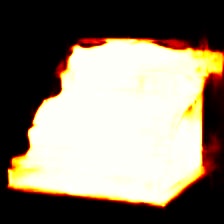}\\
     \medskip
         &  
     &  \rotatebox{90}{~~\scriptsize BASNet~\cite{qin2019basnet}} &
         \includegraphics[width=0.1\linewidth, height=80pt]{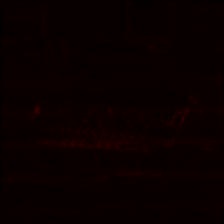} & 
  \includegraphics[width=0.1\linewidth, height=80pt]{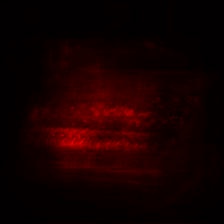} &  
    \includegraphics[width=0.1\linewidth, height=80pt]{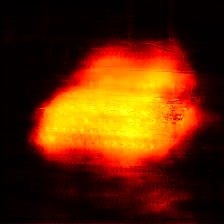} & 
        \includegraphics[width=0.1\linewidth, height=80pt]{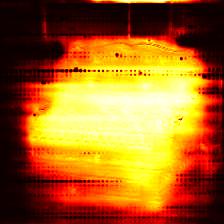} & 
        \includegraphics[width=0.1\linewidth, height=80pt]{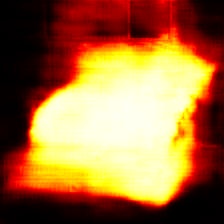} & 
    \includegraphics[width=0.1\linewidth, height=80pt]{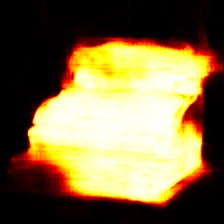} &  
\includegraphics[width=0.1\linewidth, height=80pt]{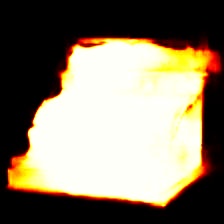}\\
\hline 
\\

    \multirow{2}{*}[20pt]{    \includegraphics[width=0.1\linewidth, height=80pt]{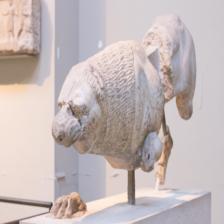}}  &  
\multirow{2}{*}[20pt]{
    \includegraphics[width=0.1\linewidth, height=80pt]{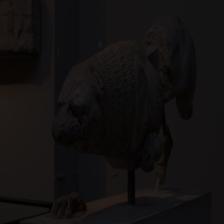}} & \rotatebox{90}{~~~\scriptsize CSNet ~\cite{gao2020highly} } & 
        \includegraphics[width=0.1\linewidth, height=80pt]{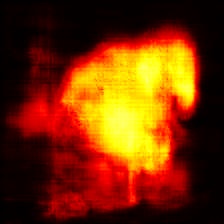} &  
      \includegraphics[width=0.1\linewidth, height=80pt]{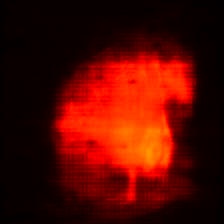} &
    \includegraphics[width=0.1\linewidth, height=80pt]{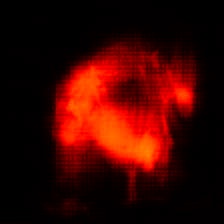} & 
        \includegraphics[width=0.1\linewidth, height=80pt]{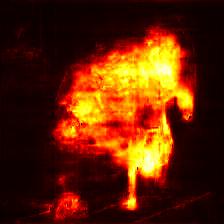} & 
    \includegraphics[width=0.1\linewidth, height=80pt]{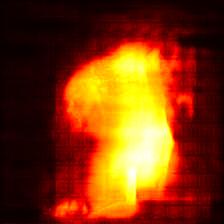}&
    \includegraphics[width=0.1\linewidth, height=80pt]{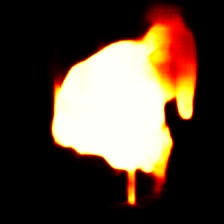} &  
    \includegraphics[width=0.1\linewidth, height=80pt]{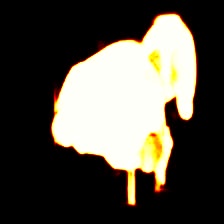}\\
         &  
     &  \rotatebox{90}{~~\scriptsize BASNet~\cite{qin2019basnet}} &
         \includegraphics[width=0.1\linewidth, height=80pt]{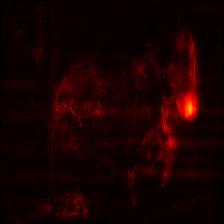} & 
\includegraphics[width=0.1\linewidth, height=80pt]{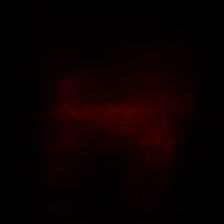} &  
    \includegraphics[width=0.1\linewidth, height=80pt]{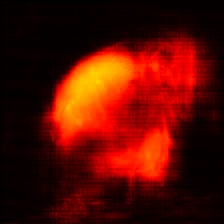} &  
    \includegraphics[width=0.1\linewidth, height=80pt]{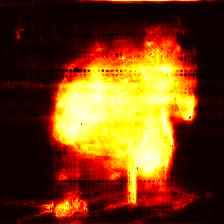} & 
        \includegraphics[width=0.1\linewidth, height=80pt]{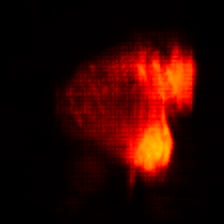} & 
        
    \includegraphics[width=0.1\linewidth, height=80pt]{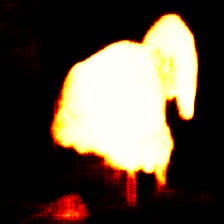} &  
\includegraphics[width=0.1\linewidth, height=80pt]{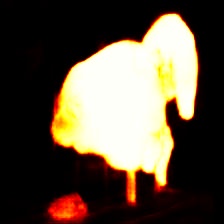}
\\
{Real} & {Real } && { \footnotesize  \textbf{a) Tr:} Well-lit} & { \footnotesize \textbf{b)Tr:}CUT} & { \footnotesize \textbf{c) Tr:}HiDT\par} &{ \footnotesize \textbf{d)Tr:}CoMoGAN\par} &{ \footnotesize\textbf{e) Tr:} FDA\par} &{\footnotesize\textbf{f) Tr:} Ours\par}  &{ \footnotesize\textbf{g) Tr:} Well-lit\par}\\
 well-lit image& low-light image&& {\footnotesize \textbf{Te:}EnLow \par}& {\footnotesize \textbf{Te:}Low-light \par}& {\footnotesize \textbf{Te:}Low-light \par} &  {\footnotesize \textbf{Te:}Low-light \par} &  {\footnotesize \textbf{Te:}Low-light \par} &  {\footnotesize \textbf{Te:}Low-light \par} &{\footnotesize \textbf{Te:}Well-lit \par}

    \end{tabular}

\caption{[\textbf{Tr}: \textit{Training}; \textbf{Te}: \textit{Testing}; \textbf{EnLow}: \textit{Enhanced low-light using Zero-DCE}~\cite{guo2020zero}] Saliency Detection by CSNet~\cite{gao2020highly} and BASNet \cite{qin2019basnet} on \textit{real} low-light images from the SICE dataset~\cite{SICE}. \textit{(a)}: Enhancing low-light images  barely improves the performance of the networks trained for well-lit images. \textit{(b), (c), (d), (e)}: Marginal improvements are observed when the networks are trained on images simulated using CUT~\cite{park2020contrastive}, HiDT~\cite{anokhin2020high}, CoMoGAN~\cite{comogan} and FDA~\cite{yang2020fda}. \textit{(f)}: 
Training models on our proxy low-light images significantly improves saliency detection on real low-light images and the predictions are close to \textit{(g)}.}
\label{Qualitatitve Analysis}
\end{figure*}

\begin{table*}[t!]
\large
\centering
\caption{Quantitative results for saliency detection  averaged over SICE’s \cite{SICE} real low-light images. The best result is in \textbf{bold} and second best is \underline{underlined}. Our proposed strategy significantly outperforms existing methods.}
\label{Quantative Analysis}
\setlength{\tabcolsep}{16pt}
\def\arraystretch{1.4705}
\begin{tabular}{c|cccccc}
\hline
\multicolumn{1}{l}{\textbf{}} &
  \textbf{\begin{tabular}[c]{@{}c@{}} CUT\\~\cite{park2020contrastive}\end{tabular}} &
  \textbf{\begin{tabular}[c]{@{}c@{}} HiDT\\~\cite{anokhin2020high}\end{tabular}} &

   \textbf{\begin{tabular}[c]{@{}c@{}} CoMoGAN\\~\cite{comogan}\end{tabular}} &
     \textbf{\begin{tabular}[c]{@{}c@{}} Zero-DCE\\~\cite{guo2020zero}\end{tabular}} &
  \textbf{\begin{tabular}[c]{@{}c@{}}FDA\\~\cite{yang2020fda}\end{tabular}} &
  \textbf{\begin{tabular}[c]{@{}c@{}}Ours\end{tabular}} \\ \hline  \vspace{-6pt}
  \\

&\multicolumn{6}{c}{\textbf{BASNet ~\cite{qin2019basnet}}}                                       \\
\textbf{E-measure$\uparrow$} & 0.391 & 0.453 &  0.423& 0.512  & \underline{0.599} & \textbf{0.602} \\
\textbf{S-measure$\uparrow$} & 0.323 & 0.344 & 0.401& 0.382 & \underline{0.568}  & \textbf{0.831} \\
\textbf{F-measure$\uparrow$} & 0.596 & 0.609 &  0.731 & 0.712 & \underline{0.874} & \textbf{0.921} \\
\textbf{MAE$\downarrow$}       & 0.462 & 0.311 & 0.243 & 0.296 & \underline{0.168} & \textbf{0.092}\\ 

&\multicolumn{6}{c}{\textbf{CSNet ~\cite{gao2020highly}}}                                        \\
\textbf{E-measure$\uparrow$} & 0.498 & 0.518 &   \underline{0.621} & 0.611 &  0.587 & \textbf{0.675} \\
\textbf{S-measure$\uparrow$} & 0.388 & 0.417 & 0.532 & 0.503 &   \underline{0.631} & \textbf{0.801} \\
\textbf{F-measure$\uparrow$} & 0.621 & 0.693 & \underline{0.756} & 0.732 &  0.755 & \textbf{0.923} \\
\textbf{MAE$\downarrow$}       & 0.321 & 0.249 & 0.221& 0.256 &  \underline{0.201} & \textbf{0.105} \\ \hline
\end{tabular}%

\end{table*}
\subsection{Low-light and well-lit fusion}
Fig.~\ref{fig:proposed_method} shows the various steps involved in our transformation pipeline. Given any real well-lit image $I_{\textrm{well}} \in \mathbb{R}^{H\times W\times 3}$, we randomly choose a real low-light image $I_{\textrm{low}}$ from a pool of real low-light images and resize it to $I_{\textrm{well}}$'s resolution. 
We next decompose the images into their respective amplitude and phase spectrums using the 2D Fourier Transform $\mathcal F$ as
\begin{eqnarray}
A^{\textrm{well}}, \measuredangle^{\textrm{well}} = \mathcal F (I_{\textrm{well}})\text{ and } A^{\textrm{low}}, \measuredangle^{\textrm{low}} = \mathcal F (I_{\textrm{low}}).
\end{eqnarray}

The image semantics are better preserved in the phase response \cite{phase_oppenheim} and so we do not modify $\measuredangle^{\textrm{well}}$.
We however, compute a weighted average of $A^{\textrm{well}}$ and $A^{\textrm{low}}$ to obtain the fused amplitude spectrum $A^{\textrm{fused}}$. For the fusion, more weightage is given to $A^{\textrm{well}}$ for high frequencies and to $A^{\textrm{low}}$ for low frequencies (See Eq.~\ref{eq:fusion}). We do this to ensure that the proxy image $I_{\textrm{prx}}$ has the semantics of $I_{\textrm{well}}$ and the style of $I_{\textrm{low}}$~\cite{image_processing_book2}.
\begin{equation}
A^{\textrm{fused}}_{m,n} = \alpha_{m,n} \cdot A^{\textrm{low}}_{m,n} + (1-\alpha_{m,n}) \cdot A^{\textrm{well}}_{m,n}
\label{eq:fusion}
\end{equation}

During fusion it is also necessary to ensure a smooth transition of frequencies, otherwise the proxy image $I_{\textrm{prx}}$ will have significant ringing artifacts due to Gibbs effect~\cite{Oppenheim_book_discrete}.
Our fusion weights $\alpha_{m,n}$ are inspired from the classical Blackman windowing~\cite{Oppenheim_book_discrete}. We empirically found that it is also necessary to retain the DC frequencies of $I_{\textrm{well}}$, otherwise the overall contrast of $I_{\textrm{prx}}$ is destroyed (see Fig. \ref{Ablation}). We therefore compute fusion over a band of frequencies and not over the entire spectrum. Formally, $\alpha_{m,n}$ is computed as
\begin{equation}
\alpha_{m,n} = \begin{cases}
          w_{m,n} & \forall \, m, n \in \mathcal{R}_{u}-\mathcal{R}_{l}\\
          0 &\text{otherwise}.
     \end{cases}
\label{eq:modified}
\end{equation}
where,
\begin{eqnarray}
        &w_{m,n}=\left[0.42+0.5\cos\left ( \frac{2\pi m}{\lambda_u \cdot H} \right )+0.08\cos\left ( \frac{4\pi m}{\lambda_u \cdot H} \right ) \right] \nonumber \\
         &\times \left[0.42+0.5\cos\left ( \frac{2\pi n}{\lambda_u \cdot W} \right )+0.08\cos\left ( \frac{4\pi n}{\lambda_u \cdot W} \right ) \right]
        \label{eq:blackman}\\
        &\mathcal{R}_{l}  \leftarrow m\in
        [-\lambda_l \frac{H}{2},\lambda_l \frac{H}{2}],\text{ and } n\in
        [-\lambda_l \frac{W}{2},\lambda_l \frac{W}{2}] \nonumber \\
        &\mathcal{R}_{u}  \leftarrow m\in
        [-\lambda_u \frac{H}{2},\lambda_u \frac{H}{2}],\text{ and } n\in
        [-\lambda_u \frac{W}{2},\lambda_u \frac{W}{2}] \nonumber \\
        & 0 \le \lambda_l < \lambda_u < 1
        \label{eq:regions}
\end{eqnarray}
Finally, $I_{\textrm{prx}}$ is obtained using the inverse Fourier transform as shown in Eq.~\ref{eq:idft}. $\gamma > 1$ controls the overall brightness of $I_{\textrm{prx}}$. Increasing the value of $\gamma$ yields a darker proxy low-light image $I_{\textrm{prx}}$.
\begin{eqnarray}
I_{\textrm{prx}} = \left[\mathcal F^{-1} (A^{\textrm{fused}},\measuredangle^{\textrm{well}})\right]^{\gamma}
\label{eq:idft}
\end{eqnarray}

Empirically, we observed that visual artifacts begin to appear as we increase the value of $\lambda_{l}$ and $\lambda_{u}$. Therefore, for our simulation, we used $\lambda_{l}=0.01$ and $\lambda_{u}=0.1$ (See Sec.~\ref{sec:ablation}).
Our proposed method can be iteratively applied to all well-lit images belonging to a dataset. For this, only few real low-light images are required for transformation. The details for transforming such well-lit datasets are given in algorithm~\ref{alg:ALG1}. Also for this algorithm to work, we do not require a paired set of well-lit and low-lit images, and they can belong to cameras of different make and model or even depict different scenes.

\begin{table}[t!]
\large
\centering
\caption{Comparison of the training time and number of parameters used by various methods to translate well lit images into proxy low-light images. Compared to other methods which have millions of parameters, FDA and our strategy contain only a couple of hyper-parameters. Thus FDA and our method do not require several hours of training time.}
\label{Time-complexity}
\def\arraystretch{1.7}
\setlength{\tabcolsep}{1pt}
\begin{tabular}{cccccc}
\hline
\multicolumn{1}{l}{\textbf{}} & \textbf{CUT} & \textbf{HiDT} &\textbf{CoMoGAN}&  \textbf{FDA} & \textbf{Ours} \\ \hline
\textbf{Parameters}           & 18.7M        & 9.8M  & 56.8M        & 1            & 2            \\
\textbf{Train Time (in hrs)}        & 24           & 24   & 48 &         N/A        & N/A          \\ \hline
\end{tabular}%
\end{table}

\begin{figure*}
  \centering
  \begin{tabular}{ c @{\hspace{2pt}} c @{\hspace{2pt}} c @{\hspace{2pt}} c @{\hspace{2pt}} c @{\hspace{2pt}}  c @{\hspace{2pt}} c @{\hspace{2pt}}}
  \multicolumn{7}{l}{\includegraphics[width=1.9\columnwidth, height=.05\columnwidth]{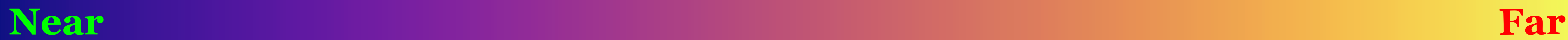}} \\
    \includegraphics[width=.31\columnwidth, height=65pt]{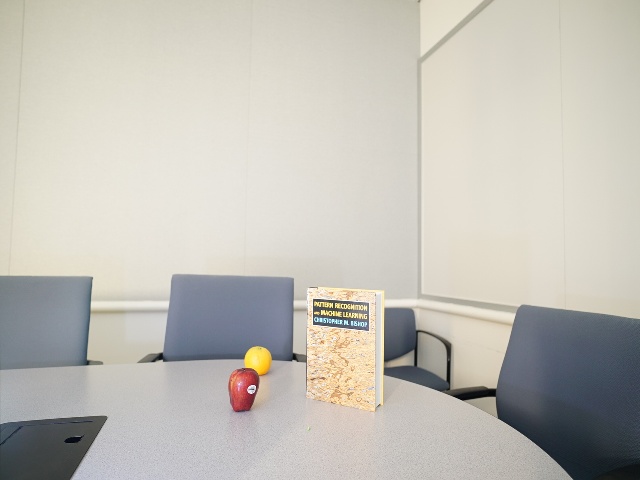} &
      \includegraphics[width=.31\columnwidth, height=65pt]{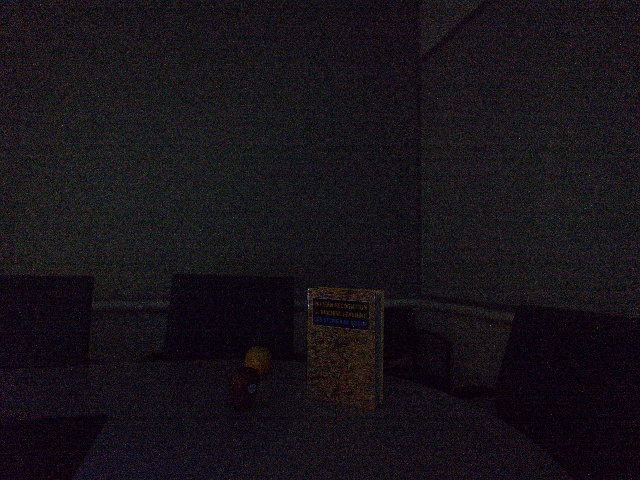}&
      \includegraphics[width=.31\columnwidth, height=65pt]{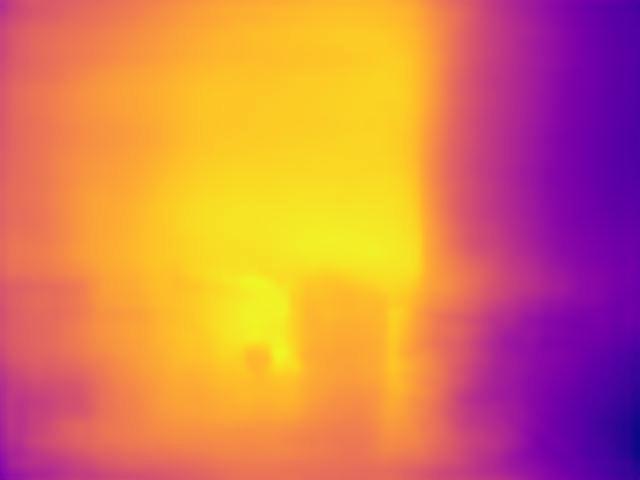}&
      \includegraphics[width=.31\columnwidth, height=65pt]{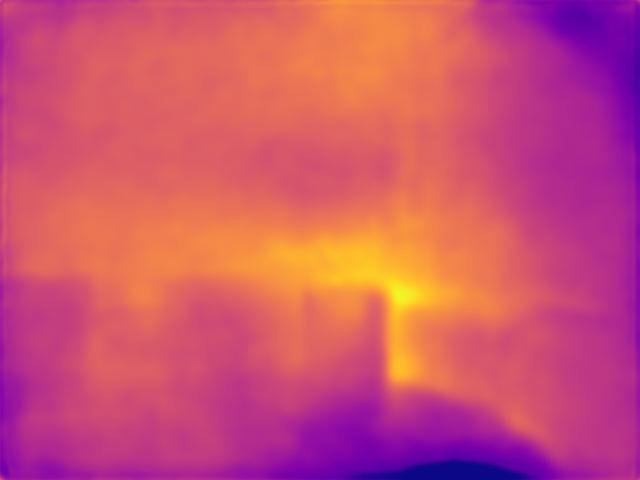}&
       \includegraphics[width=.31\columnwidth, height=65pt]{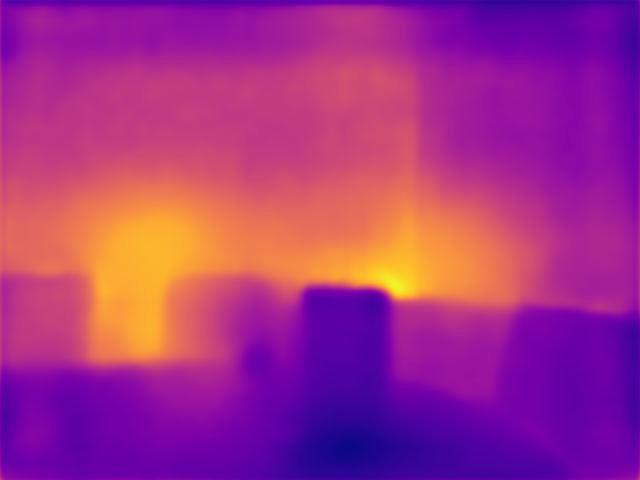}&
           \includegraphics[width=.31\columnwidth, height=65pt]{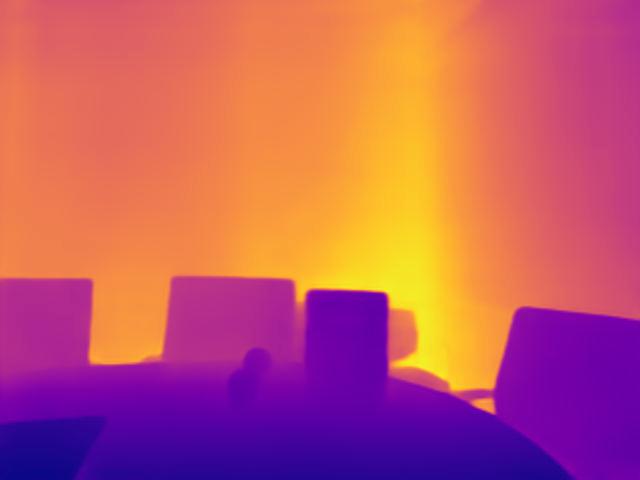} \\
    \includegraphics[width=.31\columnwidth, height=65pt]{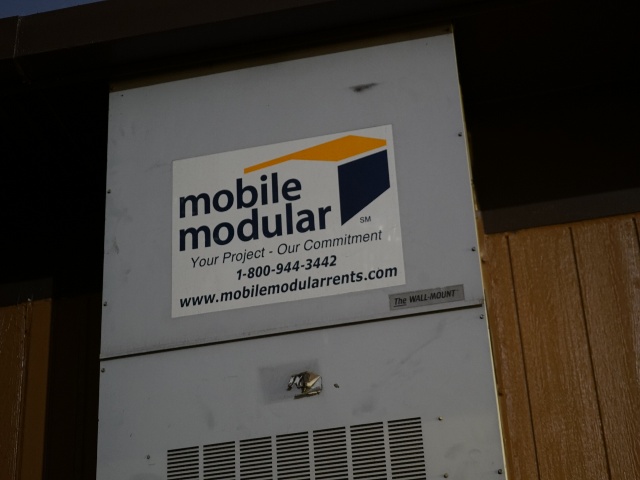} &
      \includegraphics[width=.31\columnwidth, height=65pt]{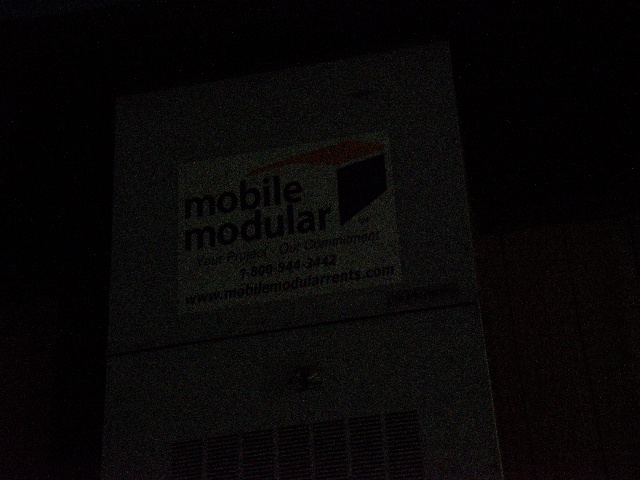}&
      \includegraphics[width=.31\columnwidth, height=65pt]{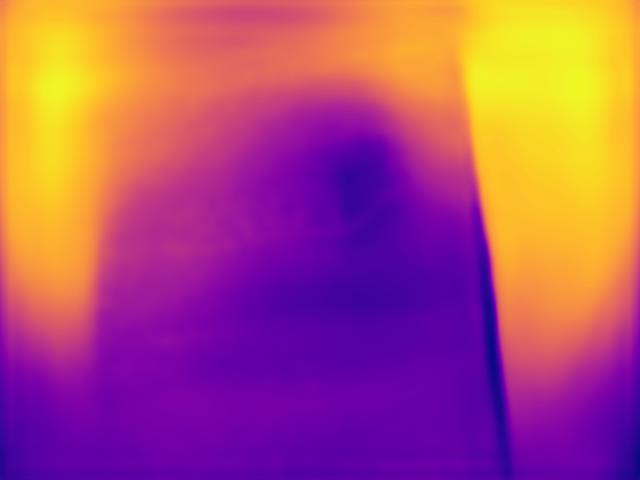}&
      \includegraphics[width=.31\columnwidth, height=65pt]{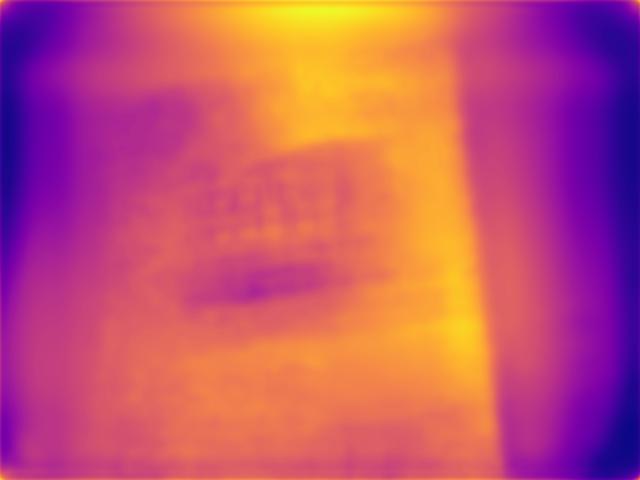}&
       \includegraphics[width=.31\columnwidth, height=65pt]{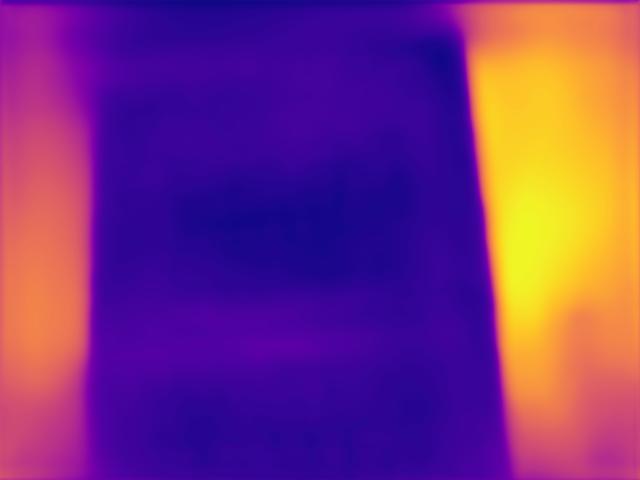}&
       \includegraphics[width=.31\columnwidth, height=65pt]{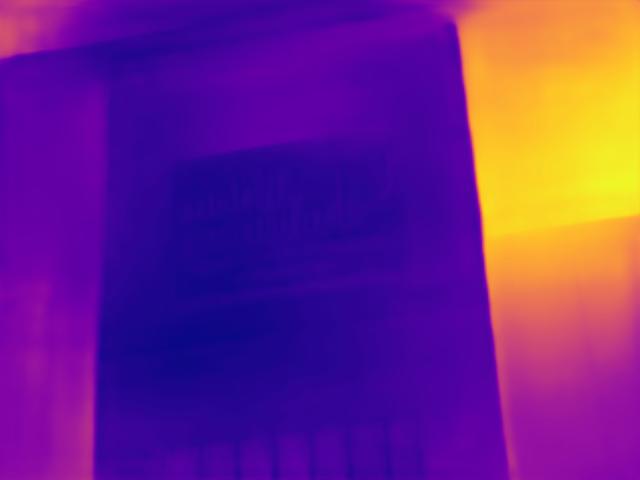}& 
       \\
    \small  Real Well-lit &    \small  Real low-light &    \small \textbf{(a) Tr:} Well-lit &     \small \textbf{(b) Tr:} FDA &   \small \textbf{(c) Tr:} Ours &   \small \textbf{(d) Tr:} Well-lit \\
    &&  \small \textbf{Te:} Real low-light & \small \textbf{Te:} Real low-light & \small \textbf{Te:} Real low-light & \small \textbf{Te:} Real Well-lit
\end{tabular}
  \caption{Depth estimation using AdaBins~\cite{adabins} on \textit{real} low-light images from the SID dataset~\cite{SID}. \textit{(a)}: AdaBins when trained on well-lit images degenerates for low-light conditions. \textit{(b)}: Training AdaBins using FDA barely improves the performance. \textit{(c)}: Training AdaBins on our proxy low-light images significantly improves depth estimation for real low-light images. Our results are close to ground truth shown in \textit{(d)}.}
  \label{Depth-Estimation-Visual}
\end{figure*}

\begin{table*}[t!]
\large
\centering
\caption{Quantitative comparison for depth estimation on real low-light images~\cite{SID}. The best result is in \textbf{bold} and second best is \underline{underlined}. Our method outperforms FDA.}
\label{depth estimation quantitative}
\def\arraystretch{1.7}
\setlength{\tabcolsep}{16pt}
\begin{tabular}{c|ccccc}
\hline
\textbf{Trained On}            & \textbf{$\delta_{1}$}$\uparrow$    & \textbf{$\delta_{2}$}$\uparrow$     & \textbf{$\delta_{3}$}$\uparrow$     & \textbf{REL}$\downarrow$    & \textbf{RMSE}$\downarrow$   \\ \hline
\textbf{Well-lit~\cite{NYU}}  & \underline{0.456}          & 0.71           & 0.878          & 0.389          & 0.725          \\
\textbf{FDA~\cite{yang2020fda}}         & 0.454    & \underline{0.794}    & \underline{0.939}   & \underline{0.318}    & \underline{0.644}    \\
\textbf{Ours}        & \textbf{0.523} & \textbf{0.833} & \textbf{0.961} & \textbf{0.276} & \textbf{0.569} \\ \hline
\end{tabular}
\end{table*}

\section{Experiments}

\subsection{Experimental Settings}
\label{sec: exp_settings}
To evaluate the proposed technique for salient object detection we use the NLPR~\cite{NLPR}, LIME~\cite{LIME}, and SICE~\cite{SICE} datasets.
The NLPR dataset contains $1000$ well-lit images of size $640 \times 480$ with corresponding GT annotations for salient objects.
LIME has $10$ real low-light images from which we used $5$ images to translate well-lit images into low-light images.
The SICE dataset contains $589$ well-lit images with corresponding real low-light images of resolutions varying from $3000 \times 2000$ to $6000 \times 4000$. Proxy low-light images generated using NLPR well-lit images are used for training 
state-of-the-art saliency detection models CSNet~\cite{gao2020highly} and BASNet~\cite{qin2019basnet}
while real low-light images of SICE dataset are reserved for testing.
Due to the absence of GT annotation for real low-light images, we consider the saliency predictions of 
BASNet and CSNet 
trained for well-lit conditions on SICE's well-lit images as the ground truth respectively.

We compare the performance of our method with HiDT~\cite{anokhin2020high}, CUT~\cite{park2020contrastive}, CoMoGAN~\cite{comogan} and FDA~\cite{yang2020fda}.
HiDT, CUT and CoMoGAN are GAN based deep learning networks for image translation, while FDA uses classical signal processing for domain adaptation.
The low-light images generated by all these methods from the well-lit NLPR dataset are then used to re-train 
BASNet and CSNet. FDA and our method uses $5$ real low-light images from the LIME dataset for low-light image conversion. CUT has to be re-trained for this task since it was not designed for well-lit to low-light transformation. As $5$ images are too less for training GAN based models, additional $3000$ images from the Ex-Dark dataset~\cite{ExDark} are used when training GAN based models. We also tried fine-tuning HiDT and CoMoGAN, but as they are specifically designed for low-light translation, the performance of pre-trained models is better and we use them for all comparisons.

We additionally compare with Zero-DCE~\cite{guo2020zero} which is used to enhance low-light images as a pre-processing step. We could not compare with works of Xu \textit{et al.}~\cite{xu2020exploring},~\cite{mu2019salient},~\cite{xu2018extended} since neither their code nor their dataset is publicly available.

We use PyTorch running on a CPU with $32$GB RAM and a $12$GB K80 GPU for implementing the proposed method. Unless stated otherwise, lower-frequency~($\lambda_{l}$), upper-frequency~($\lambda_{u}$) and gamma~($\gamma$) are set to $0.01, 0.10$ and $3.5$, respectively. 
Other parameters such as the loss function, optimiser and data augmentations are as mentioned in the available codes of above stated methods.

\begin{figure*}
  \centering
  \begin{tabular}{ c @{\hspace{5pt}} c @{\hspace{5pt}} c @{\hspace{5pt}} c @{\hspace{2pt}} c @{\hspace{2pt}}  c @{\hspace{2pt}}}
  \multicolumn{6}{c}{\includegraphics[width=1.95\columnwidth, height=.04\columnwidth]{IEEE-SPL-LaTex-1/images/colorbar_hot_new.pdf}} \\
    \includegraphics[trim={ 0 0 90pt 10pt},clip,width=.38\columnwidth, height=70pt]{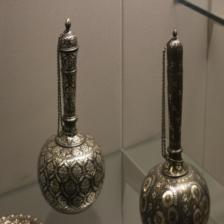} &
    \includegraphics[trim={ 0 0 90pt 10pt},clip,width=.38\columnwidth, height=70pt]{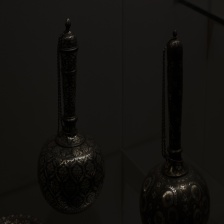} &    

      \includegraphics[trim={ 0 0 90pt 10pt},clip,width=.38\columnwidth, height=70pt]{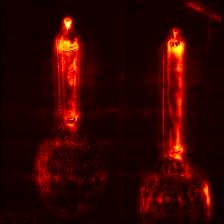}&
      \includegraphics[trim={ 0 0 90pt 10pt},clip,width=.38\columnwidth, height=70pt]{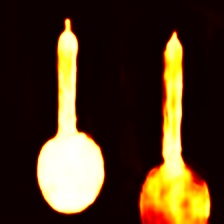}&
       \includegraphics[trim= {0 0 90pt 10pt}, clip,width=.38\columnwidth, height=70pt]{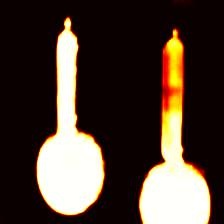}&
\\
    \small  Real Well-lit &    \small  Real low-light &    \small \textbf{(a) Tr:} Real low-light &     \small \textbf{(b) Tr:} Ours  &   \small \textbf{(c) Tr:} Ours+fine-tuned \\
    
    &&  \small \textbf{Te:} Real low-light & \small \textbf{Te:} Real low-light & \small on real low-light images \\
    &&&&  \textbf{Te:} \small Real low-light \\
\end{tabular}
  \caption{Qualitative comparison of saliency maps generated on \textit{real} low-light images from the SICE dataset when CSNet is trained on: \textit{(a)} real low-light images, \textit{(b)} our proxy low-light images, \textit{(c)} fine-tuning (b) on real low-light images. Without using our synthetic images, it is not possible to get good performance under low-light conditions because of the absence of publicly available large-scale datasets for low-light saliency detection.
  }
  \label{fig: Fine-tuned Model}
\end{figure*}

\begin{table*}[t!]
\large
\centering
\caption{Quantitative comparison for CSNet trained on: (a) real low-light images from SICE, (b) our proxy images and (c) our proxy images followed by fine tuning on real low-light images from SICE. The best result is in \textbf{bold} and second best is \underline{underlined}. Training CSNet on real low-light images yields poor results due to the absence of large-scale datasets for low-light saliency detection.
However, using our synthetic images to increase the training size  significantly improves performance as indicated in columns $2$ and $3$.}
\label{table: finetune}
\def\arraystretch{1.7}
\setlength{\tabcolsep}{16pt}
\begin{tabular}{c|cccc}
\hline
\textbf{Trained On}&\textbf{Real low-light images}&\textbf{Ours} & \textbf{Ours+fine-tuned on real low-light images}\\ \hline
\textbf{S-measure}$\uparrow$ & 0.619 & \underline{0.801} & \textbf{0.821} \\
\textbf{F-measure}$\uparrow$ & 0.823 & \underline{0.923} & \textbf{0.939} \\\hline
\end{tabular}
\end{table*}

\subsection{Qualitative and Quantitative comparisons}
In Fig.~\ref{Qualitatitve Analysis} we visually compare the saliency maps generated by BASNet and CSNet in different situations.
We observe that the simple pre-processing step of enhancing low-light images using Zero-DCE before feeding them to BASNet~\cite{qin2019basnet} and CSNet~\cite{gao2020highly} trained on well-lit images yields unsatisfactory results. 
Marginal improvements are observed if well-lit images are first translated to low-light images using HiDT~\cite{anokhin2020high}, CUT~\cite{park2020contrastive} and CoMoGAN~\cite{comogan} and then used to re-train 
BASNet and CSNet.
This is mainly because, adversarial training is often susceptible to training instabilities and unnatural artifacts in the generated images.
Training using FDA proxy images yields better predictions compared to other methods, but is still quite inferior to ground truth. This is because, as discussed in Sec.~\ref{sec:ablation}, FDA transformed images have considerable ringing artifacts.
Predictions using our transformation not only outperform all existing methods but are almost at par with ground truth.
Our superiority is also supported by Table.~\ref{Quantative Analysis} where we outperform existing methods on all four metrics, namely, E-Measure~\cite{E-measure}, S-measure~\cite{S-measure}, F-measure~\cite{F-measure} and Mean-Absolute-Error (MAE).

\subsection{Time-Complexity}
\label{sec:time}
Table.~\ref{Time-complexity} reports the training time required by CUT, HiDT, CoMoGAN, FDA and the proposed method for generating proxy low-light images.
This includes the time needed for training GAN based methods. We see that GAN based methods take at least $48 \times$ more time than FDA and Ours to transform images. 
Compared to deep learning networks, which have millions of learnable parameters, the proposed transformation has only $2$ hyper-parameters i.e., $\lambda_{l}$ and $\lambda_{u}$.
FDA has only one hyper-parameter, $\beta$, which is comparable to $\lambda_{u}$ in our algorithm. 
If $\gamma$ is also considered, hyper-parameter count for FDA and ours increase by one. Thus, our method not only exhibits qualitative and quantitative superiority but is also fast with a low number of parameters.

\subsection{Generalizabilty}
Our method is easy to generalize to other computer vision tasks. We demonstrate this by extending our pipeline for depth estimation under extreme low-light conditions. Specifically, we re-train a recent depth estimation network AdaBins~\cite{adabins} on our proxy low-light images generated using well lit images present in the NYU dataset~\cite{NYU} and then test it on real extreme low-light images from the SID dataset~\cite{SID}.
The NYU dataset consists of $640 \times 480$ well-lit images with ground truth depth annotations and the SID dataset consists of $4256 \times 2848$ real night-time images with their corresponding well-lit images. For this experiment we use only the low-light images captured with $0.1$s exposure.
For transforming NYU well-lit images we used just \textit{one} real low-light image from the SID dataset with lower-frequency~($\lambda_{l}$), upper-frequency~($\lambda_{u}$) and gamma~($\gamma$) set to $0.01, 0.1$ and $6$ respectively. We have increased the $\gamma$ from $3.5$ to $6$ as SID images are much more dark than SICE dataset. Similar settings are used for the FDA pipeline.
For benchmarking, we compute GT depth by passing the well-lit SID images through the original AdaBins trained for well-lit images.
The qualitative results can be found in Fig.~\ref{Depth-Estimation-Visual} and quantitative results in Table.~\ref{depth estimation quantitative} where we use the same metrics as used in the AdaBins paper.

\begin{figure*}
  \centering
  \Large
  \begin{tabular}{ c @{\hspace{2pt}} c @{\hspace{2pt}} c @{\hspace{2pt}} c @{\hspace{2pt}} c @{\hspace{2pt}}  @{\hspace{2pt}} c @{\hspace{2pt}}}
     \includegraphics[width=.175\linewidth]{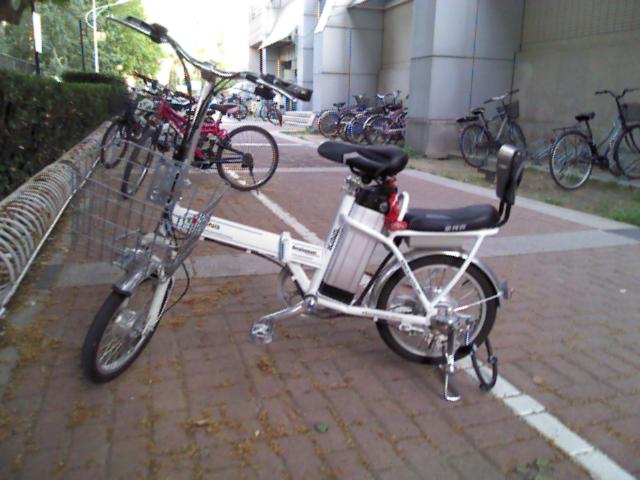} &
    \includegraphics[width=.175\linewidth ]{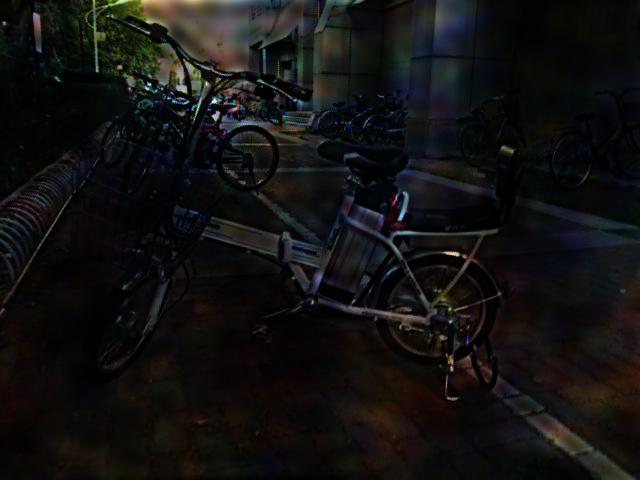} &
     \includegraphics[width=.175\linewidth ]{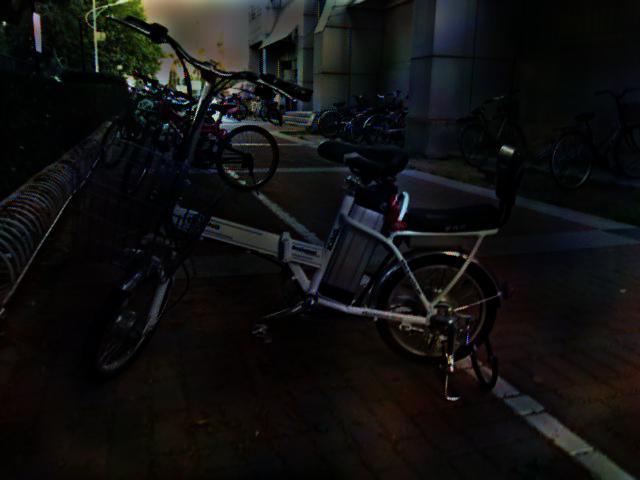} &
      \includegraphics[width=.175\linewidth ]{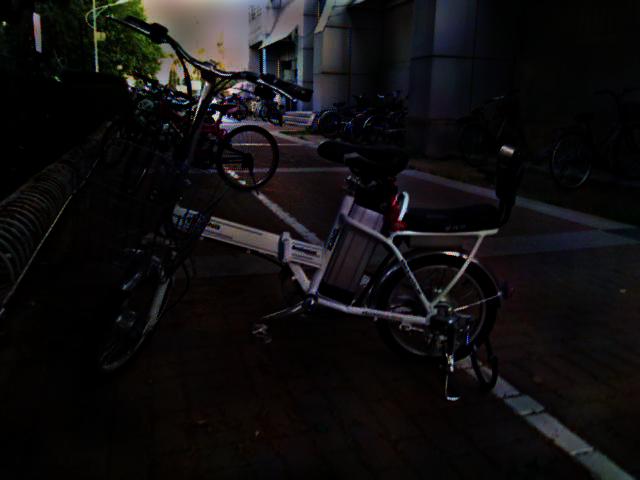}&
    \includegraphics[width=.175\linewidth ]{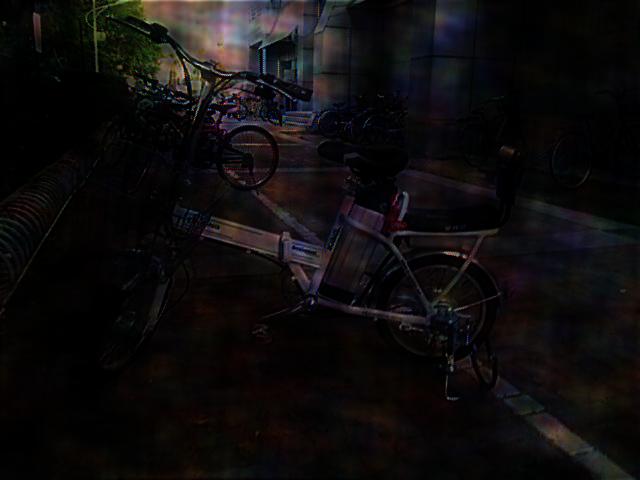}\\
    \includegraphics[width=.175\linewidth]{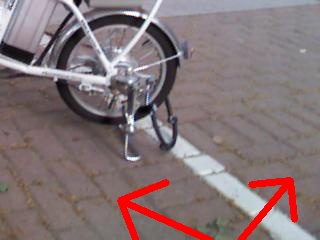} &
    \includegraphics[width=.175\linewidth ]{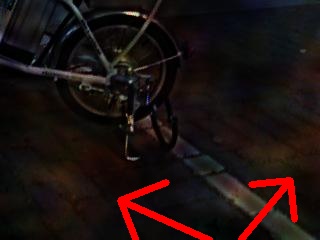} &
     \includegraphics[width=.175\linewidth ]{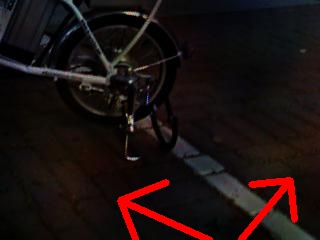} &
      \includegraphics[width=.175\linewidth ]{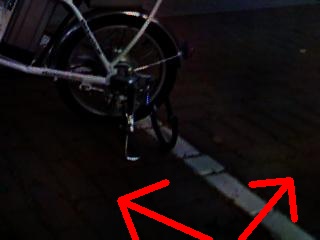}&
    \includegraphics[width=.175\linewidth ]{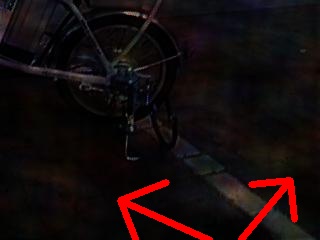}
    
    \\
    
 \footnotesize Well-lit image & \footnotesize $\textbf{i) }\lambda_{l}=0$ &\footnotesize $\textbf{ii) }\lambda_{l}=0$ & \footnotesize $\textbf{iii) }\lambda_{l}=0.01$ & \footnotesize $\textbf{iv) }\lambda_{l}=0.01$\\
 &\footnotesize $\lambda_{u}=0.1$ &\footnotesize $\lambda_{u}=0.1$ & \footnotesize $\lambda_{u}=0.1$ & \footnotesize $\lambda_{u}=0.5$\\
 &\footnotesize $w=1$&\footnotesize $w=$ Eq.~\ref{eq:blackman} &\footnotesize$w=$ Eq.~\ref{eq:blackman} &\footnotesize$w=$ Eq.~\ref{eq:blackman}\\
 &&&\footnotesize{(Proposed)} &
  \end{tabular}
\caption{Ablation study showing the effect of $\lambda_{l}$, $\lambda_{u}$ and $w$ in generating $I_{\textrm{prx}}$. $\gamma$ was set to $2.5$ for all the images.
Color and ringing artifacts can be observed in \textbf{i)}. However, our windowing technique suppresses these ringing artifacts as shown in \textbf{ii)}. But, color artifacts are still present in \textbf{ii)} which are indicated by the red arrows. These color artifacts are diminished by using our proposed band-pass filtering instead of low-pass filtering as shown in \textbf{iii)}. Using a large value of $\lambda_{u}$ degrades the visual quality as shown in \textbf{iv)}.}

\label{Ablation}
\end{figure*}

\subsection{Training on real low-light images}
\label{fine_tuning experiments}
There is no publicly available large scale dataset to train networks for low-light saliency detection. We however show that such networks can be first trained on our proxy images and then fine-tuned on a limited number of real low-light images to improve performance. We do this by evaluating the performance of CSNet under three scenarios: (i) training on a limited number of real low-light images from the SICE~\cite{SICE} dataset, (ii) training on our proxy image dataset obtained from the well-lit NLPR saliency dataset which has large number of images and (iii) by fine-tuning the network obtained in (ii) using limited number of real low-light images from (i).

The NLPR dataset consists of well-lit images with corresponding ground truth saliency maps but lacks low-light images.
On the other hand, the SICE dataset has well-lit and low-light pairs but lacks ground truth saliency maps.
Thus as described in Sec.~\ref{sec: exp_settings}, for (i) we treated the saliency maps generated by passing well-lit SICE images through CSNet trained for well-lit conditions as the ground truth. After discarding the images for which the ground truth maps were not appropriate by manual inspection, we finally obtained $156$ real low-light images with ground truth saliency.
For (ii) we translated well-lit NLPR images into proxy low-light images while retaining original saliency ground truth (see Sec.~\ref{sec: exp_settings} for details).

Table.~\ref{table: finetune} and Fig.~\ref{fig: Fine-tuned Model}  respectively present the quantitative and qualitative results for the different scenarios. The poor performance of the network in Fig.\ref{fig: Fine-tuned Model}\textit{(a)} is due to the limited number of real low-light images available for training. However, using our proxy images for pre-training and then fine-tuning with these limited number of real low-light images (in our case $156$) boosts the network's performance as shown in Fig.\ref{fig: Fine-tuned Model}\textit{(d)}.

\subsection{Ablation Studies}
\label{sec:ablation}
Fig.~\ref{Ablation} shows the ablation studies conducted on our method by choosing well-lit images from the NLPR dataset and a real low-light image from the SID~\cite{SID} dataset.
In Fig.~\ref{Ablation} i) we do not use weighted averaging for fusion and instead in Eq.~\ref{eq:blackman} we set $w = 1$ which causes sharp discontinuities at the cut-off frequencies $\frac{\lambda_{u}H}{2}$ and $\frac{\lambda_{u}W}{2}$.
We additionally do not retain the DC frequencies of $I_{\textrm{well}}$ by setting $\lambda_{l}=0$.
Clearly, the transformed images lack contrast and exhibit severe ringing artifacts.
Except for the $\gamma$ correction, Fig.~\ref{Ablation} i) is same as FDA.
In Fig.~\ref{Ablation} ii) we enforce a smooth fusion of well-lit and low-light images by using $w$ as defined in Eq.~\ref{eq:blackman}. This helps limit the Gibbs phenomenon leading to removal of ringing artifacts visible in Fig.~\ref{Ablation} i).
The colors in Fig.~\ref{Ablation} ii), however, continue to be poor. For example in the second row in Fig.~\ref{Ablation} ii), the color of the road as indicated by the red arrow has reddish-brown patches. 
In Fig.~\ref{Ablation} iii) we use band-pass filtering instead of low-pass filtering by slightly increasing $\lambda_{l}$ from $0$ to $0.01$. Clearly band-pass filtering leads to better color restorations.
Finally in Fig.~\ref{Ablation} iv) we use a large value of $\lambda_{u}$ which consequently degrades the semantics of $I_{\textrm{well}}$ in the generated proxy low-light image. This is expected because a large value of $\lambda_{u}$ implies that even the high frequencies of real low-light image, which mostly capture the semantics of low-light image, are fused into the frequency spectrum of well-lit image. We, however, only wish to incorporate the style of low-light images and not their semantics into the well-lit images. 
As Fig.~\ref{Ablation} iii) qualitatively yields better low-light proxy images, we fix $\lambda_{l}$ and $\lambda_{u}$ to $0.01$ and $0.1$ respectively.

\section{Conclusion}
Existing saliency detection datasets mostly consist of well-lit images which make models trained on these datasets unsuitable for saliency detection under low-light conditions. Alleviating this problem generally involves using GAN based models which are computationally expensive and difficult to train.
We thus proposed a classical computer vision method to generate proxy low-light images from well-lit images which can be used to train models for saliency estimation under real low-light conditions. We used band-pass filtering in the Fourier domain for translating well-lit images into proxy low-light images. During filtering, we ensured a smooth fusion of frequencies which suppressed the ringing artifacts.
Our method has only a few hyper-parameters and is thus easy to generalize for different computer vision applications such as depth estimation. Specifically, we showed that models trained on our proxy low-light images outperformed existing low-light image translation methods for saliency and depth estimation under real low-light conditions.
\begin{acks}
This work was supported in part by IITM Pravartak Technologies Foundation.
\end{acks}
\bibliographystyle{ACM-Reference-Format}
\bibliography{book1}

\end{document}